# Privacy-preserving Federated Bayesian Learning of a Generative Model for Imbalanced Classification of Clinical Data

Seok-Ju Hahn[1] · *Junghye Lee[1]

Master Student     Assistant Professor

[1]School of Management Engineering, Ulsan National Institute of Science and Technology (UNIST),

Ulsan 689–798, Republic of Korea.

{sjhahn11512, junghyelee}@unist.ac.kr

**Abstract**

In clinical research, the lack of events of interest often necessitates imbalanced learning. One approach to resolve this obstacle is data integration or sharing, but due to privacy concerns neither is practical. Therefore, there is an increasing demand for a platform on which an analysis can be performed in a federated environment while maintaining privacy. However, it is quite challenging to develop a federated learning algorithm that can address both privacy-preserving and class imbalanced issues. In this study, we introduce a federated generative model learning platform for generating samples in a data-distributed environment while preserving privacy. We specifically propose approximate Bayesian computation-based Gaussians Mixture Model called 'Federated ABC-GMM', which can oversample data in a minor class by estimating the posterior distribution of model parameters across institutions in a privacy-preserving manner. PhysioNet2012, a dataset for prediction of mortality of patients in an Intensive Care Unit (ICU), was used to verify the performance of the proposed method. Experimental results proved that our method boosts classifier's performance up to nearly an ideal situation in terms of F1 score. It is believed that the proposed method can be a novel alternative to solving class imbalance problems.

## 1. Introduction

In the era of data abundance, developments of powerful machine learning algorithms have been contributing to richer analyses on many fields, thereby creating value conducive to humankind at an explosive rate [1]. This has created an impact on various areas including a clinical field, where supporting decision making of clinicians and discovering useful knowledge is now viable thanks to data and analytics [2, 3]. Especially, electronic health records (EHRs) has established in the clinical field, and their frequency of the usage has also been increasing [4]. EHRs are organized collection of data on patients' health information in a digital format. EHRs cover most events that may occur in a clinical process; medication history, prescriptions, diagnostic reports, lab tests, and images such as CT and X-ray, and vital sign are included [5]. With the advent of EHRs, high quality care for patients based on data analysis became available, thereby a medical practice, which previously heavily relied on the knowledge of physicians or clinical professionals, has been slanted toward an evidence-based approach, mining values from data [6].

However, still many challenges exist in the clinical field, which make data analysis difficult. Examples include data privacy issues such as leakage from malicious attacks [7], distributed data across multiple institutions (i.e. hospitals) [8] for some reasons, e.g. patient's hospital movement, and imbalanced data arising from lack of events [9, 10]. These aforementioned problems are frequently encountered in clinical data analysis, and a few studies have been actively conducted with the focus to each of them [11-14].

Most of the data generated and collected in hospitals often contain patients' personal information. Therefore, de-identification is essential for the use of data. Nonetheless, when non-identified data is shared between different sources, a risk of privacy leakage still exists [16]. For these reasons, most hospitals are reluctant to share data with each other. However, it is inevitable to pool horizontally distributed data when it is required to do an analysis on rare events or cases like predicting occurrence of rare diseases. In fact, an aggregation of the horizontally distributed data in one place can be realized after going through formalities like receiving an approval of Institutional Review Board (IRB), and data usage agreement. But the process is time-consuming and is not robust to privacy attack, albeit de-identification [15]. Therefore, a federated learning is being offered as a promising solution to do analysis on distributed data, not explicitly sharing or transferring data to other places [17].

A federated learning is mainly divided into two categories for vertically and horizontally distributed data. The former one is a case where samples are overlapped, but in different feature spaces. The latter one is a sample-based federated learning, of which participants have data with the same feature space (Yang et al., 2019). What this study focused on is a horizontally distributed environment. One typical method of a specific predictive model's learning in such an environment is usually done through exchanging local model's encrypted gradient update at the central server (CS) [18]. For encryption, diverse techniques are applied e.g. differential privacy [19], homomorphic encryption [20].

In clinical data analysis, there are often class imbalanced issues that make pattern recognition difficult. It is considered as an imbalanced dataset if heterogeneity exists in the distributions between classes. The problem of learning data mining algorithms with an imbalanced dataset is defined as imbalanced learning (He and Garcia, 2009), and the problematic situation comes from various settings. Many approaches have been being suggested to resolve the issue: sampling methods (oversampling [22-24] / undersampling [25-26]), cost-sensitive method [27], kernel-based methods [28], or a combination of these [29, 30].

Data from less frequent or rarely collected classes (e.g. rare diseases such as Crohn's disease) exacerbate the results of its classification. One typical way of resolving this problem is to use oversampling techniques (i.e. synthesizing minor samples). Oversampling based on adjacent sample distance (SMOTE [22], ADASYN [23]), or fitting a generative model such as a Gaussian Mixture Model (GMM) [31], Generative Adversarial Networks (GAN) [32] on the distribution of the minor samples for generating new plausible samples are often alleviating the imbalanced issue [33]. The generative model is then utilized to complement lack of data in a minor class. Those augmentation methods sometimes remedy a classifier to have better generalization power of discriminating different distributions of heterogeneous classes [34, 35].

This study proposes a federated learning framework that enables the Bayesian estimation of posterior distribution of a set of parameters of a generative model (particularly GMM in this study), to alleviate problems of imbalanced and horizontally distributed data. Such a process can be done by indirectly reflecting information of data in separate locations while protecting privacy information, not sharing data. The proposed method will be applied and validated on the real data.

## 2. Preliminaries

### 2.1. Approximate Bayesian Computation (ABC)

ABC is one of the representative methods of likelihood-free inference and a Bayesian parameter estimation method utilized in the situation where the likelihood of the model is intractable or even not defined [36]. This estimation method only requires a parametrized generative model so that data simulation is feasible when a set of parameters is given [37].

#### 2.1.1. ABC Rejection Sampling

One of the simplest ABC parameter estimation methods is ABC rejection sampling [38].

| Algorithm ABC REJECTION SAMPLING |
|---|
| Input <br> $\mathbf{x}$ : observed data, <br> $S(\cdot)$ : appropriate summary statistic, <br> $p(\boldsymbol{\theta})$ : prior distribution(s) of parameter(s) $\boldsymbol{\theta}$, <br> $\|\cdot\|$ : a discrepancy measure, <br> $\epsilon$ : sample acceptance threshold <br> $L$ : maximum number of accepted parameters. <br><br> Output <br> $p_\epsilon(\boldsymbol{\theta}\|S(\mathbf{x}))$ : a posterior distribution of the (set of) parameter(s) $\boldsymbol{\theta}$ |
| 1  *#params* = 0 |
| 2  **while** (*#params* < *L*) **do** |
| 3      Simulate (set of) parameter(s) $\boldsymbol{\theta}^*$ |
| 4          from the prior distribution $\boldsymbol{\theta}^* \sim p(\boldsymbol{\theta})$ |
| 5      Generate data $\mathbf{x}^* \sim p(\mathbf{x}\|\boldsymbol{\theta}^*)$ |
| 6      Compute $S(\mathbf{x}^*)$ |
| 7      **if** $\|S(\mathbf{x}^*) - S(\mathbf{x})\| < \epsilon$ **then** |
| 8          Store $\boldsymbol{\theta}^*$ |
| 9          *#params* += 1 |
| 10 **end** |

Generated samples are from $p_\epsilon(\boldsymbol{\theta}|S(\mathbf{x}))$ in which

$$p_\epsilon(\boldsymbol{\theta}|S(\mathbf{x})) \propto \int_{\mathcal{X}} p(\mathbf{x}^*|\boldsymbol{\theta}^*)p(\boldsymbol{\theta}^*)\mathbf{I}_{X_{\epsilon,\mathbf{x}}(\mathbf{x}^*)}\,\mathrm{d}\mathbf{x}^*$$

where

$$X_{\epsilon,\mathbf{x}}(\mathbf{x}^*) = \{\mathbf{x}^* \in \mathcal{X} \mid \|S(\mathbf{x}^*) - S(\mathbf{x})\| < \epsilon\}.$$

### 2.1.2. Summary Statistics

When $S(\cdot)$ is a summary statistic for model parameter $\boldsymbol{\theta}$, it is guaranteed that the approximated posterior distribution is equivalent to the posterior distribution estimated using original data [37].

$$p_\epsilon(\boldsymbol{\theta}|S(\mathbf{x})) = p_\epsilon(\boldsymbol{\theta}|\mathbf{x})$$

This trait emphasizes the importance of choosing a proper summary statistic of the data.

### 2.1.3. Discrepancy Metric

The discrepancy metric $\|\cdot\|$ is usually Euclidean distance when the summary statistic yields sufficiently small dimension, however, for high-dimensional data, using the Euclidean cannot guarantee to give a good result. Instead, using different discrepancy(distance) metrices such as Interquartile range, Kullback-Leibler (KL) divergence [39], or aggregating different distance measures [37] can be an alternative so that the effect of one distorted measure be mitigated by others works well.

### 2.1.4. Threshold

Finally, for the ABC rejection sampling to be converged in distribution, selection of an appropriate threshold value $\epsilon$ is essential:

$$\lim_{\epsilon \to 0} p_\epsilon(\boldsymbol{\theta}|\mathbf{x}) = p(\boldsymbol{\theta}|\mathbf{x}), \lim_{\epsilon \to \infty} p_\epsilon(\boldsymbol{\theta}|\mathbf{x}) = p(\boldsymbol{\theta})$$

For an arbitrarily large threshold, the algorithm learns nothing from the data, whereas a tiny threshold may require almost infinite computation time [37].

In summary, the posterior distribution for a parameter can be estimated without defining an explicit likelihood, and the posterior distribution obtained by ABC is for the set of parameter values whose simulated data most closely match the observed data.

## 2.2. GMM

A GMM, is a parametrized probabilistic function for estimating an unknown arbitrary probability density, assuming that all data points are from the finite mixture of Gaussian distributions. In other words, the unknown density function can be represented as the weighted sum of Gaussian components.

Suppose we have an input matrix $\mathbf{X} \in \mathbb{R}^{N \times D}$, which row and column correspond to a sample vector $\mathbf{x}_i, i = 1, \dots, N$ and a feature vector $\mathbf{x}_p, p = 1, \dots, D$, respectively, and the input data are from the mixture of $K$ Gaussian distributions. A GMM typically consists of three parameters, $\boldsymbol{\theta} = \{\pi_k, \boldsymbol{\mu}_k, \boldsymbol{\Sigma}_k\}_{k=1}^{K}$; $\pi_k$ is the responsibility value that indicates the probability that the input data belong to the k-th latent cluster; the latent cluster can be modeled by a latent random variable $\mathbf{z}$, of which distribution is parameterized by $\boldsymbol{\pi}$. The input data assigned in any latent cluster $k$ follow the normal distribution with mean $\boldsymbol{\mu}_k$ (centered value of each latent cluster) and covariance $\boldsymbol{\Sigma}_k$ (modeling the variance and co-variance of data in a specific latent cluster). Then, the density of input data through the GMM is estimated as:

$$p(\mathbf{x}) = \sum_{\mathbf{z}} p(\mathbf{z})p(\mathbf{x}|\mathbf{z}) = \sum_{k=1}^{K} \pi_k \mathcal{N}(\mathbf{x}|\boldsymbol{\mu}_k, \boldsymbol{\Sigma}_k)$$

where

$$p(\mathbf{z}) = \prod_{k=1}^{K} \pi_k^{z_k}, p(\mathbf{x}|\mathbf{z}) = \prod_{k=1}^{K} \mathcal{N}(\mathbf{x}|\boldsymbol{\mu}_k, \boldsymbol{\Sigma}_k)^{z_k}$$

$$z_k \in \{0, 1\}, \sum_k z_k = 1,$$

$$0 \le \pi_k \le 1, \sum_k \pi_k = 1,$$

$$p(z_k = 1) = \pi_k.$$

All parameters can be estimated expectation maximization (EM) algorithm [40].

However, using ABC to estimate posterior distributions of parameters, it is essential to define suitable prior distributions of each parameter [41].

For $\boldsymbol{\pi} = \{\pi_k | \pi_k \in \mathbb{R}, k = 1, \dots, K\}$, a proper prior distribution is known as a Dirichlet distribution, where $\boldsymbol{\alpha} = (\alpha_1, \alpha_2, \dots, \alpha_K)$. When $K$ is determined, the prior distribution of $\boldsymbol{\pi}$, regards $K$ as the number of categories; $\boldsymbol{\pi} \sim \text{Dir}(\boldsymbol{\pi}|\boldsymbol{\alpha})$.

Next, for $\boldsymbol{\mu} = \{\boldsymbol{\mu}_k | \boldsymbol{\mu}_k \in \mathbb{R}^{D \times 1}, k = 1, \dots, K\}$, $\boldsymbol{\Sigma} = \{\boldsymbol{\Sigma}_k | \boldsymbol{\Sigma}_k \in \mathbb{R}^{D \times D}, k = 1, \dots, K\}$, their prior distributions are intertwined; so-called Normal-Inverse-Wishart distribution. At first, a covariance matrix $\boldsymbol{\Sigma}$ is sampled from Inverse-Wishart prior distribution, parametrized by a scale matrix $\boldsymbol{\Psi} \in \mathbb{R}^{D \times D}$ which should be positive definite, and a degree of freedom, $v > D - 1$; $\boldsymbol{\Sigma} \sim \mathcal{W}^{-1}(\boldsymbol{\Sigma}|v, \boldsymbol{\Psi})$.

Then, using the sampled covariance matrix $\boldsymbol{\Sigma}$, the prior distribution of $\boldsymbol{\mu}$, a multivariate-normal distribution, becomes available: with another parameter, a location vector $\boldsymbol{m} \in \mathbb{R}^D$, and a positive real number $\kappa$; $\boldsymbol{\mu} \sim \mathcal{N}\left(\boldsymbol{\mu}|\boldsymbol{m}, \frac{1}{\kappa}\boldsymbol{\Sigma}\right)$.

In summary, each prior distribution of parameters is:

$$\boldsymbol{\pi} \sim \text{Dir}(\boldsymbol{\pi}|\boldsymbol{\alpha})$$

$$(\boldsymbol{\mu}, \boldsymbol{\Sigma}) \sim \text{NIW}(\boldsymbol{\mu}, \boldsymbol{\Sigma} | \upsilon, \boldsymbol{\Psi}, \boldsymbol{m}, \kappa)$$
$$= \mathcal{N}\left(\boldsymbol{\mu} | \boldsymbol{m}, \frac{1}{\kappa}\boldsymbol{\Sigma}\right) \mathcal{W}^{-1}(\boldsymbol{\Sigma}|\upsilon, \boldsymbol{\Psi}).$$

The graphical expression of Bayesian GMM is shown in Figure 1.

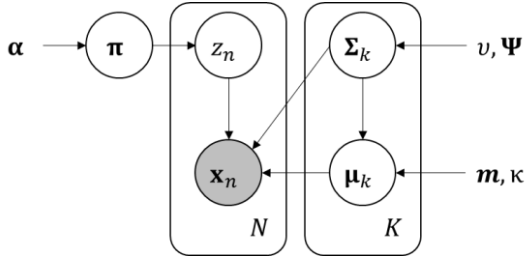

Figure 1. Graphical Model of Bayesian GMM

In this research, the posterior distribution of each parameter is estimated under the ABC framework.

## 3. Methods

### 3.1. Autoencoder (AE) as a Summary Statistic Generator

An AE [42] is a type of a neural network learned in an unsupervised manner by minimizing the reconstruction error in rebuilding the encoded output into the original input. A basic AE is composed of two symmetric sub-networks: an encoder network and a decoder network. The former is to shrink the input vector into a lower dimensional vector, and the latter is to reconstruct the original input from compressed vector. The AE is mainly used to derive latent representations by reducing the dimensions of the input feature space while removing noise inherent in the data. The objective function of AE is usually provided as:

$$\text{minimize } \|\mathbf{x} - \mathbf{x}'\|_2^2$$

, where the reconstructed input is

$$\mathbf{x}' = h_{dec}\left(\mathbf{W}_{dec}\left(h_{enc}(\mathbf{W}_{enc}\mathbf{x} + \mathbf{b}_{enc})\right) + \mathbf{b}_{dec}\right),$$

and $h(\cdot), \mathbf{W}, \mathbf{b}$ denote an element-wise activation function, a weight matrix, a bias vector of encoder and decoder, respectively.

In this research, the AE is utilized as a summary statistic generator for ABC parameter estimation. Since a deeper neural network architecture is advantageous (Goodfellow et al., 2016), more than a single layer is used to assemble the encoder and decoder networks. While achieving its original purpose, the AE at the same time optimizes compressed representations to be well-classified by logistic regression. By inserting a bypath to encoded output nodes and adding a new objective function, the modified AE called *moAE* is able to generate a low dimensional vector that can be well separated by a linear model (Figure 2).

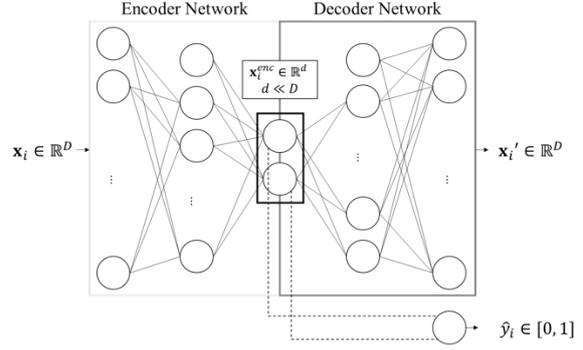

Figure 2. Illustration of *moAE*

The objective function of *moAE* can be defined as:

$$\text{minimize } \alpha \sum_{i=1}^{N} \|\mathbf{x}_i - \mathbf{x}'_i\|_2^2$$
$$-\frac{\beta}{N} \sum_{i=1}^{N} \left(y_i \ln \sigma(\mathbf{w}_{clf}^T \mathbf{x}_i^{enc})\right.$$
$$\left. + (1 - y_i) \ln\left(1 - \sigma(\mathbf{w}_{clf}^T \mathbf{x}_i^{enc})\right)\right)$$

, where $\alpha$ and $\beta$ are arbitrary small positive numbers for adjusting the contribution of each loss function and $\sigma(\cdot)$ is a sigmoid function; $\mathbf{x}_i$, $\mathbf{x}'_i$, and $\mathbf{x}_i^{enc}$ represent the $i$-th original vector, reconstructed vector, and latent vector respectively, and the corresponding $y_i \in \{0,1\}$ indicates one of true labels.

Hyperbolic tangent function (tanh) is used in the layer right before nodes generating latent representation to make data from different sites be compressed and laid on the similar range, [-1, 1]. Identity function (linear activation function) and sigmoid activation function are used on output nodes for reconstruction and on an output node for classification, each. For other nodes in hidden layers, Scaled-exponential Linear Unit, SeLU [44], is used as an activation function. For optimizing the objective function, Adam [45] update rule is adopted.

It should be noted that the latent representations generated from the *moAE* cannot be recovered to their original input if each site does not disclose their trained decoder networks i.e. weights to the public, thereby data can be preserved [46].

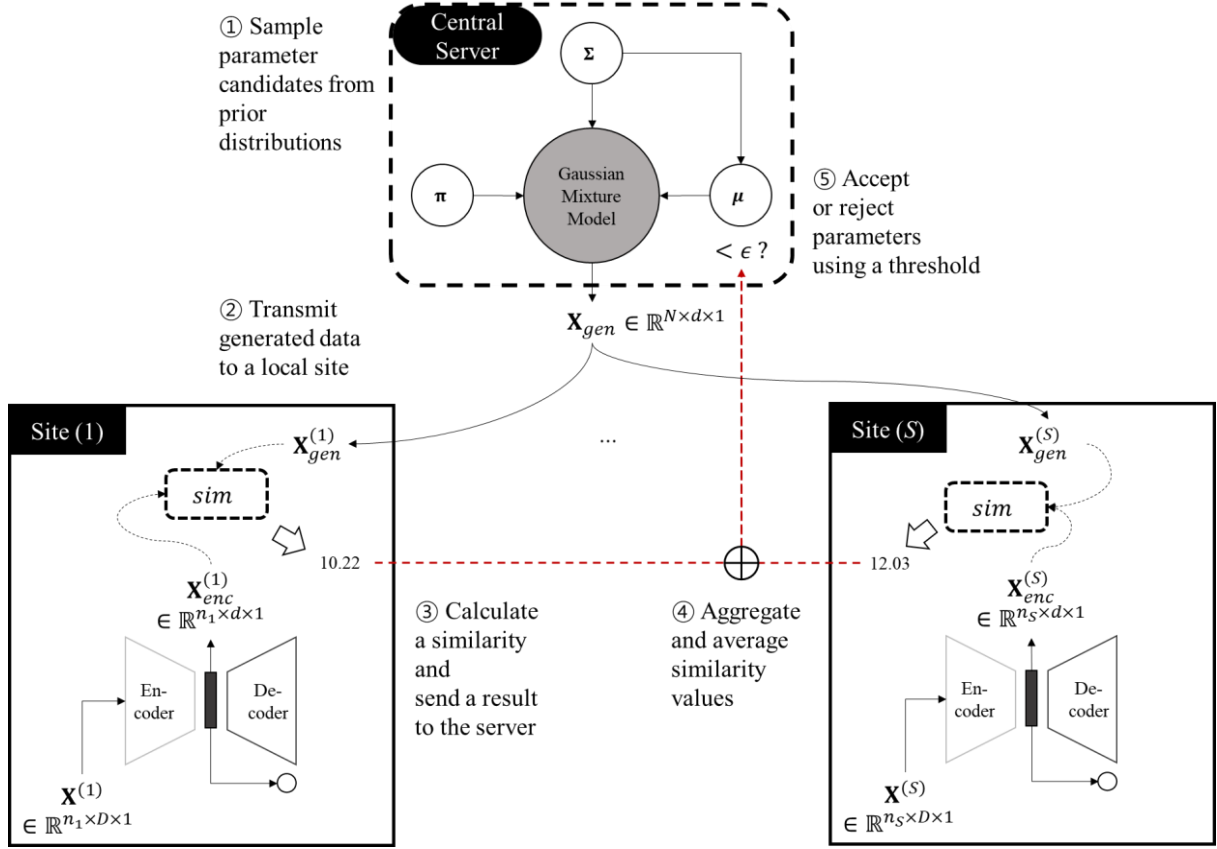

Figure 3. Overall Framework of FEDERATED ABC-GMM

## 3.2. Bayesian Inference of GMM in a distributed environment using ABC (Federated ABC-GMM)

The proposed framework is illustrated in Figure 3. First, the CS estimates the posterior distributions of parameters of GMM, by exchanging information with local sites. The information is a similarity between generated samples, which are aimed to be close to perturbed samples, and modified samples in each site via *moAE*. The main assumption of the framework is that all participants in the framework have data of equivalent features, i.e. horizontally distributed data. Besides, each site is reluctant to share their local data even if the data is perturbed via the *moAE*. Therefore, the only information transmitted from local sites to the CS is the discrepancy calculated in each local site. The discrepancy between generated data, of which dimensions are equivalent to dimensions of latent vectors in each site, and local data is measured with proper metrics. In this study, the following combination of Euclidean distance and KL divergence is used as a similarity metric:

$$sim(\mathbf{X}_{enc}^{total}, \mathbf{X}_{gen}^{total})$$
$$= \sum_{i=1}^{S} euclidean(\mathbf{X}_{enc}^{(i)}, \mathbf{X}_{gen}^{(i)}) + D_{KL}(\mathbf{X}_{enc}^{(i)} || \mathbf{X}_{gen}^{(i)})$$

, where

$$euclidean(\mathbf{X}_{enc}^{(i)}, \mathbf{X}_{gen}^{(i)}) = \sum_{j=1}^{n_i} \|\mathbf{x}_{enc,j}^{(i)} - \mathbf{x}_{gen,j}^{(i)}\|_2^2,$$

$$D_{KL}(\mathbf{X}_{enc}^{(i)}, \mathbf{X}_{gen}^{(i)}) = \sum P(\mathbf{x}_{enc}^{(i)}) \log \frac{P(\mathbf{x}_{enc}^{(i)})}{P(\mathbf{x}_{gen}^{(i)})},$$

for $S$ sites, $\mathbf{X}_{enc}^{total}, \mathbf{X}_{gen}^{total} \in \mathbb{R}^{N \times d \times 1}$ are aggregated tensors having $n_i$ numbers of encoded/generated tensors $\mathbf{X}_{enc}^{(i)}, \mathbf{X}_{gen}^{(i)} \in \mathbb{R}^{n_i \times d \times 1}$, $\sum_i n_i = N$. Each of which has $d$ dimensional vectors $\mathbf{x}_{enc}^{(i)}, \mathbf{x}_{gen}^{(i)}$.

Based on the pre-set threshold value, a sampled parameter candidate from each prior distribution is determined to be accepted or rejected at the CS. When the process is iterated, the accepted set of parameters becomes close to its true posterior distribution.

Once the inference to the posterior distributions is completed, plausible perturbed samples in the minor class can be generated at the CS, and then such samples, transmitted from the CS, can boost the classification accuracy for the imbalanced data at the local site. in summary, the framework with the appropriate similarity measure enables to train model containing global information without observing any local data in raw. The inference process of FEDERATED ABC-GMM is developed on the Engine for Likelihood Free Inference (ELFI) [47] framework.

| **Algorithm** FEDERATED ABC-GMM |   |
|---|---|
| | Input |
| | $\mathbf{X}_{train}^{(i)} \in \mathbb{R}^{N \times D \times 1}$: training data at site $i$, |
| | $\mathbf{y}_{train}^{(i)} \in \mathbb{R}^{N \times 1}$ : corresponding label at site $i$, |
| | $moAE^{(i)}(\cdot)$ : Modified AE at site $i$ |
| |    (returns encoder network $Enc^{(i)}(\cdot)$, |
| |    used to generate latent representations), |
| | #epochs : number of epochs to train the $moAE$, |
| | $sim(\cdot)$ : discrepancy metric defined in 4.2 |
| | $\epsilon$ : a proper threshold value |
| | $Dir(\boldsymbol{\pi}|\boldsymbol{\alpha})$ : prior of $\boldsymbol{\pi}$ |
| | $NIW(\boldsymbol{\mu}, \boldsymbol{\Sigma}|v, \boldsymbol{\Psi}, \boldsymbol{m}, \kappa)$ : prior of $\boldsymbol{\mu}, \boldsymbol{\Sigma}$, |
| | $L$ : maximum number of accepted parameters, |
| | $S$ : number of sites, |
| | $d$ : latent dimension. |
| | |
| | Output |
| | $p_\epsilon\left(\boldsymbol{\theta}\middle|Enc(\mathbf{X}_{train}^{total})\right)$ : posterior distributions |
| |    of the set of parameters $\boldsymbol{\theta} = \{\boldsymbol{\pi}, \boldsymbol{\mu}, \boldsymbol{\Sigma}\}$. |
| 1 | *(At each local site)* |
| 2 | **for** $i$ in $S$ **do** |
| 3 |   **for** _ in #epochs **do** |
| 4 |     Train $moAE^{(i)}(\mathbf{X}_{train}^{(i)}, \mathbf{X}_{train}^{(i)}, \mathbf{y}_{train}^{(i)})$ |
| 5 |     Generate latent representations |
| 6 |     $\mathbf{X}_{enc}^{(i)} = Enc^{(i)}(\mathbf{X}_{train}^{(i)}) \in \mathbb{R}^{n_i \times d \times 1}$. |
| 7 | **end** |
| 8 | #params = 0 |
| 9 | **while** (#params < $L$) **do** |
| 10 | *(At CS)* |
| 11 |   Simulate set of parameters |
| 12 |     $\boldsymbol{\theta}^* = \{\boldsymbol{\pi}^*, \boldsymbol{\mu}^*, \boldsymbol{\Sigma}^*\}$ from each prior |
| 13 |     $\boldsymbol{\pi}^* \sim Dir(\boldsymbol{\pi}|\boldsymbol{\alpha})$, |
| 14 |     $(\boldsymbol{\mu}^*, \boldsymbol{\Sigma}^*) \sim NIW(\boldsymbol{\mu}, \boldsymbol{\Sigma}|v, \boldsymbol{\Psi}, \boldsymbol{m}, \kappa)$. |
| 15 |   Generate data $\mathbf{X}_{gen} \sim p(\mathbf{X}|\boldsymbol{\theta}^*)$. |
| 16 |   Split $\mathbf{X}_{gen}$ according to $n_i$, |
| 17 |     $\mathbf{X}_{gen}^{(i)} \in \mathbb{R}^{n_i \times d \times 1}$. |
| 18 |   Send $\mathbf{X}_{gen}^{(i)}$ to each site $i$. |
| 19 | *(At each local site)* |
| 20 |   Calculate $\phi^{(i)} = sim(\mathbf{X}_{enc}^{(i)}, \mathbf{X}_{gen}^{(i)})$. |
| 21 |   Send a value $\phi^{(i)}$ to the CS. |
| 22 | *(At CS)* |
| 23 |   Receive and collect the value $\phi^{(i)}$ |
| 24 |   Set $\bar{\phi} = \frac{1}{S}\sum_{i=1}^{S}\phi^{(i)}$. |
| 25 |   **if** $\bar{\phi} < \epsilon$ **then** |
| 26 |     Store $\boldsymbol{\theta}^* = \{\boldsymbol{\pi}^*, \boldsymbol{\mu}^*, \boldsymbol{\Sigma}^*\}$. |
| 27 |     #params += 1 |
| 28 | **end** |

## 4. Experiments and Results

### 4.1. Simulation of Horizontally Distributed Data

We used PhysioNet2012 to simulate horizontally distributed data environment. PhysioNet2012 is a dataset of The PhysioNet Computing in Cardiology Challenge 2012, which aimed to predict mortality in intensive care unit [48]. Among 4,000 samples in a training set, 310 samples satisfying two assumptions are selected (i.e. $N = 310$): (1) imbalanced data, (2) data which classified well when all data is available at one place, but is not when only partial data is provided.

88 out of 189 features are remained (i.e. $D = 88$), where removed ones have Pearson's R correlation coefficient bigger than 0.8. Three independent sites (i.e. $S = 3$) are simulated by splitting data into three without duplication. At each site, each partitioned dataset is again divided into training and test sets in a stratified manner to keep class ratio.

Table 1. Description of Horizontally Distributed Setting

|  | Site 1 | Site 2 | Site 3 |
|---|---|---|---|
| Training data with label 0 (Major Class) | 51 | 46 | 62 |
| Training data with label 1 (Minor Class) | 9 | 8 | 10 |
| Class Ratio (Major:Minor) | 5.67:1 | 5.75:1 | 6.20:1 |

Before training an AE at each site, we need to do standardization on each training data first for faster convergence (Ioeffe and Szegedy, 2015) of Adam optimizer. Using the calculated mean and covariance of the training data, test data is also standardized as follows.

$$\tilde{\mathbf{x}}_{train,j} = \frac{\mathbf{x}_{train,j} - \mu_{train,j}}{\sigma_{train,j}},$$

$$\tilde{\mathbf{x}}_{test,j} = \frac{\mathbf{x}_{test,j} - \mu_{train,j}}{\sigma_{train,j}},$$

$$j = 1, \dots, D.$$

The summary statistic of each training data can be obtained from each trained encoder network. Table 2 shows details of the AE structure used in this study.

The encoding dimensions $d$ (i.e. dimensions of a latent vector) can be determined as an arbitrary number satisfying $d \ll D$. In this experiment, $d$ is set to be 24. Finally, for learning a GMM at the CS, the number of components should be pre-determined, and we set the floor of 90% of minor samples as the latent cluster number, i.e. $K = floor\big((9 + 8 + 10) * 0.9\big) = 24$.

Table 2. Summary of *moAE* Structure

| Name | Layer | Output Dimension | # Parameters. (with bias) | Activation Function | Connection |
|---|---|---|---|---|---|
| **Encoder** | input | 88 | 0 | - | encoded_1 |
| | encode_1 | 64 | 5696 | SeLU | encoded_2 |
| | encode_2 | 32 | 2080 | SeLU | latent |
| **Latent Nodes** | latent | 24 | 792 | tanh | decoded_1, logistic_regression |
| **Decoder** | decode_1 | 32 | 800 | SeLU | decoded_2 |
| | decode_2 | 64 | 2112 | SeLU | recon |
| | recon | 88 | 5720 | linear | - |
| **Classifier** | logistic_regression | 1 | 25 | sigmoid | - |

*4.2. Experimental Results*

To validate the proposed framework, several tests were done on test datasets of each local site. As logistic regression is included during the training process of local AEs, the performance of logistic regression is used as a validation criterion in the experiment. Other measures were also used in the experiment, even though focusing on F1 score is reasonable as the original problem is to predict the patient mortality, which is class imbalanced. Therefore, all results are based on the cut-off value that maximizes the F1 score under the training set:

$$F1 = \frac{2 \times Precision \times Recall}{Precision + Recall}.$$

As one of the main assumptions is that the classifier learned on the global training dataset (i.e. global classifier) outperforms the classifier learned on the local training dataset (i.e. local classifier). At first, test datasets from each local site were used to evaluate the performance of the global classifier (denoted by 'Global Site $i$'); the results are deemed to be an upper bound of non-global classifiers.

Then, the performances of two local classifiers was measured at each site: once from the original local training set (denoted as 'Site $i$ Raw') and one from the augmented training set ('Site $i$ OS'), which includes both original local data of the minor class and oversampled data generated from the local GMM fit with the local data of the minor class.

Finally, FEDERATED ABC-GMM is learned on the minority class training data of each site, and then samples the scarce data; such samples are expected to contain global information. Then, the performance of the proposed method was estimated (denoted as 'Site $i$ ABC').

All oversampling methods in the experiment are forced to increase samples of the minor class to the same number as the major class. Results are shown in Table 3.

## 5. Discussion

We demonstrated the feasibility of Distributed ABC-GMM; in terms of F1 score, our proposed framework outperforms local classifier's and even approaches the global classifier. However, there are still several limitations in this study.

The experiments were conducted on a single machine (with different processes) to serve as a proof-of-concept. In practice, it is required to deploy the algorithm in multiple machines.

There is an intrinsic limitation of the ABC estimation method, since it only depends on indirect information that a discrepancy measure yields [37], which causes the gap between our method and the global classifier. To overcome this drawback, a sequential Monte-Carlo ABC (SMC-ABC), which is an efficient and accurate sampling strategy with the high convergence speed [50], can be an alternative to improve the ABC estimation part. Moreover, there has to be a complementary approach to help a GMM generate samples guaranteed since there is no guarantee that all samples generated by the GMM will always help to enhance the performance of the local classifier. Therefore, we can think of re-filtering plausible and helpful samples, e.g. based on the nearest-neighbor approach [51], to ensure a certain level of performance.

Our framework should be tested on unstructured data like images such as X-ray and signals like vital signs. Plus, for high-dimensional, or big sample size data, the proposed method is cost-expensive since it relies mainly on a sampling approach [52]. Therefore, more efficient sampling methods like aforementioned SMC-ABC, and Hamiltonian ABC [53] would be used in a future work.

Table 3. Classification results on each local site

| | Global Site 1 | Site 1 Raw | Site 1 OS | Site 1 ABC | Global Site 2 | Site 2 Raw | Site 2 OS | Site 2 ABC | Global Site 3 | Site 3 Raw | Site 3 OS | Site 3 ABC |
|---|---|---|---|---|---|---|---|---|---|---|---|---|
| Accuracy | 0.8250 | 0.7750 | 0.8000 | 0.6250 | 0.8056 | 0.8333 | 0.8333 | 0.7778 | 0.8125 | 0.8333 | 0.8333 | 0.7917 |
| Sensitivity | 0.5000 | 0.1667 | 0.1667 | 0.8333 | 0.6000 | 0.4000 | 0.4000 | 0.6000 | 0.5000 | 0.1667 | 0.1667 | 0.5000 |
| Specificity | 0.8824 | 0.8824 | 0.9118 | 0.5882 | 0.8387 | 0.9032 | 0.9032 | 0.8065 | 0.8571 | 0.9286 | 0.9286 | 0.8333 |
| Precision | 0.4286* | 0.2000 | 0.2500 | **0.2632** | 0.3750 | 0.4000* | 0.4000* | 0.3333 | 0.3333* | 0.2500 | 0.2500 | **0.3000** |
| Recall | 0.5000* | 0.1667 | 0.1667 | **0.8333** | 0.6000* | 0.4000 | 0.4000 | **0.6000** | 0.5000* | 0.1667 | 0.1667 | **0.5000** |
| F1 | 0.4615* | 0.1818 | 0.2000 | **0.4000** | 0.4615* | 0.4000 | 0.4000 | **0.4286** | 0.4000* | 0.2000 | 0.2000 | **0.3750** |
| Threshold | - | - | - | 8 | - | - | - | 8 | - | - | - | 8 |
| Cut-off | 0.3743 | 0.8453 | 0.9473 | 0.4311 | 0.3743 | 0.8391 | 0.9356 | 0.5051 | 0.3743 | 0.8877 | 0.9374 | 0.5041 |

As in typical Bayesian estimation methods, appropriate prior distributions should be provided for successful likelihood-free inference of parameters. In the case of GMM, well-defined priors are provided in the literature, but this is not always the case. In addition, it is almost impossible to use ABC in non-parametric models with too many parameters such as an artificial neural network, because ABC is mostly applicable to parametric generative models.

Nevertheless, it is still attractive that absorbing information with no data shared between organizations is possible under the ABC framework. Combining other federated learning frameworks with the method proposed in this research in a complementary manner would promote secure and efficient data analysis on a distributed clinical dataset.

## 6. Conclusion

This study proposed a privacy-preserving federated Bayesian learning framework. To avoid privacy leakage while achieving the original purpose, a classification of imbalanced data, *moAE* is suggested. Plus, a method of learning a generative model for oversampling data in a minor class using ABC, FEDERATED ABC-GMM, is proposed. The ABC method also enhances privacy protection by handling perturbed data, while federating information from inaccessible area through similarity metric. Experiments conducted on real clinical dataset validated that the proposed methods are applicable in a data-distributed environment in reality.

## References

1. Obermeyer, Z., & Emanuel, E. J. (2016). Predicting the future—big data, machine learning, and clinical medicine. The New England journal of medicine, 375(13), 1216.

2. Cosgriff, C. V., Celi, L. A., & Sauer, C. M. (2018). Boosting clinical decision-making: machine learning for intensive care unit discharge.

3. Beam, A. L., & Kohane, I. S. (2018). Big data and machine learning in health care. Jama, 319(13), 1317-1318.

4. Rose, S. (2018). Machine learning for prediction in electronic health data. JAMA network open, 1(4), e181404-e181404.

5. Yadav, P., Steinbach, M., Kumar, V., & Simon, G. (2018). Mining electronic health records (EHRs): a survey. ACM Computing Surveys (CSUR), 50(6), 85.

6. Agrawal, R., Delen, D., & Benjamin, D. (2019). Clinical Intervention Research with EHR: A Big Data Analytics Approach.

7. Abouelmehdi, K., Beni-Hessane, A., & Khaloufi, H. (2018). Big healthcare data: preserving security and privacy. Journal of Big Data, 5(1), 1.

8. Rosales, R. E., & Rao, R. B. (2010). Guest Editorial: Special Issue on impacting patient care by mining medical data. Data mining and knowledge discovery, 20(3), 325-327.

9. Wu, J., Roy, J., & Stewart, W. F. (2010). Prediction modeling using EHR data: challenges, strategies, and a comparison of machine learning approaches. Medical care, S106-S113.

10. Manogaran, G., Chilamkurti, N., & Hsu, C. H. (2018). Emerging trends, issues, and challenges in Internet of Medical Things and wireless networks. Personal and Ubiquitous Computing, 22(5-6), 879-882.

11. Liu, X., Lu, R., Ma, J., Chen, L., & Qin, B. (2015). Privacy-preserving patient-centric clinical decision

support system on naive Bayesian classification. IEEE journal of biomedical and health informatics, 20(2), 655-668.

12. Gutierrez, O., Saavedra, J. J., Zurbaran, M., Salazar, A., & Wightman, P. M. (2018, October). User-Centered Differential Privacy Mechanisms for Electronic Medical Records. In 2018 International Carnahan Conference on Security Technology (ICCST) (pp. 1-5). IEEE.

13. Raisaro, J. L., Troncoso-Pastoriza, J., Misbach, M., Sousa, J. S., Pradervand, S., Missiaglia, E., ... & Hubaux, J. P. (2018). Medco: Enabling secure and privacy-preserving exploration of distributed clinical and genomic data. IEEE/ACM transactions on computational biology and bioinformatics.

14. Huang, L., Shea, A. L., Qian, H., Masurkar, A., Deng, H., & Liu, D. (2019). Patient clustering improves efficiency of federated machine learning to predict mortality and hospital stay time using distributed electronic medical records. Journal of Biomedical Informatics, 103291.

15. El Emam, K., & Dankar, F. (2012). U.S. Patent No. 8,316,054. Washington, DC: U.S. Patent and Trademark Office.

16. Culnane, C., Rubinstein, B. I., & Teague, V. (2017). Health data in an open world. arXiv preprint arXiv:1712.05627.

17. Hartmann, F. (2018). Federated Learning.

18. Yang, Q., Liu, Y., Chen, T., & Tong, Y. (2019). Federated machine learning: Concept and applications. ACM Transactions on Intelligent Systems and Technology (TIST), 10(2), 12.

19. Dwork, C. (2011). Differential privacy. Encyclopedia of Cryptography and Security, 338-340.

20. Takabi, H., Hesamifard, E., & Ghasemi, M. (2016, December). Privacy preserving multi-party machine learning with homomorphic encryption. In 29th Annual Conference on Neural Information Processing Systems (NIPS).

21. He, H., & Garcia, E. A. (2008). Learning from imbalanced data. IEEE Transactions on Knowledge & Data Engineering, (9), 1263-1284.

22. Chawla, N. V., Bowyer, K. W., Hall, L. O., & Kegelmeyer, W. P. (2002). SMOTE: synthetic minority over-sampling technique. Journal of artificial intelligence research, 16, 321-357.

23. He, H., Bai, Y., Garcia, E. A., & Li, S. (2008, June). ADASYN: Adaptive synthetic sampling approach for imbalanced learning. In 2008 IEEE International Joint Conference on Neural Networks (IEEE World Congress on Computational Intelligence) (pp. 1322-1328). IEEE.

24. Lunardon, N., Menardi, G., & Torelli, N. (2014). ROSE: A Package for Binary Imbalanced Learning. R journal, 6(1).

25. Anand, A., Pugalenthi, G., Fogel, G. B., & Suganthan, P. N. (2010). An approach for classification of highly imbalanced data using weighting and undersampling. Amino acids, 39(5), 1385-1391.

26. Yen, S. J., & Lee, Y. S. (2009). Cluster-based under-sampling approaches for imbalanced data distributions. Expert Systems with Applications, 36(3), 5718-5727.

27. Liu, X. Y., & Zhou, Z. H. (2006, December). The influence of class imbalance on cost-sensitive learning: An empirical study. In Sixth International Conference on Data Mining (ICDM'06) (pp. 970-974). IEEE.

28. Hong, X., Chen, S., & Harris, C. J. (2007). A kernel-based two-class classifier for imbalanced data sets. IEEE Transactions on neural networks, 18(1), 28-41.

29. Mathew, J., Luo, M., Pang, C. K., & Chan, H. L. (2015, November). Kernel-based SMOTE for SVM classification of imbalanced datasets. In IECON 2015-41st Annual Conference of the IEEE Industrial Electronics Society (pp. 001127-001132). IEEE.

30. Wang, S., Li, Z., Chao, W., & Cao, Q. (2012, June). Applying adaptive over-sampling technique based on data density and cost-sensitive SVM to imbalanced learning. In The 2012 International Joint Conference on Neural Networks (IJCNN) (pp. 1-8). IEEE.

31. Duda, R. O., & Hart, P. E. (1973). Pattern classification and scene analysis. Wiley, New York, 81, 269-296.

32. Goodfellow, I., Pouget-Abadie, J., Mirza, M., Xu, B., Warde-Farley, D., Ozair, S., ... & Bengio, Y. (2014). Generative adversarial nets. In Advances in neural information processing systems (pp. 2672-2680).

33. Douzas, G., & Bacao, F. (2018). Effective data generation for imbalanced learning using conditional generative adversarial networks. Expert Systems with applications, 91, 464-471.

34. Liu, A., Ghosh, J., & Martin, C. E. (2007, June). Generative Oversampling for Mining Imbalanced Datasets. In DMIN (pp. 66-72).

35. Yin, H. L., & Leong, T. Y. (2010). A model driven approach to imbalanced data sampling in medical decision making. In MedInfo (pp. 856-860).

36. Marin, J. M., Pudlo, P., Robert, C. P., & Ryder, R. J. (2012). Approximate Bayesian computational methods. Statistics and Computing, 22(6), 1167-1180.

37. Turner, B. M., & Van Zandt, T. (2012). A tutorial on approximate Bayesian computation. Journal of Mathematical Psychology, 56(2), 69-85.

38. Pritchard, J. K., Seielstad, M. T., Perez-Lezaun, A., & Feldman, M. W. (1999). Population growth of human Y chromosomes: a study of Y chromosome microsatellites. Molecular biology and evolution, 16(12), 1791-1798.

39. Jiang, B. (2018, March). Approximate Bayesian computation with Kullback-Leibler divergence as data discrepancy. In International Conference on Artificial Intelligence and Statistics (pp. 1711-1721).

40. Dempster, A. P., Laird, N. M., & Rubin, D. B. (1977). Maximum likelihood from incomplete data via the EM algorithm. Journal of the Royal Statistical Society: Series B (Methodological), 39(1), 1-22.

41. Bishop, C. M. (2006). Pattern recognition and machine learning. springer.

42. Rumelhart, D. E., Hinton, G. E., & Williams, R. J. (1985). Learning internal representations by error propagation (No. ICS-8506). California Univ San Diego La Jolla Inst for Cognitive Science.

43. Goodfellow, I., Bengio, Y., & Courville, A. (2016). Deep learning. MIT press.

44. Klambauer, G., Unterthiner, T., Mayr, A., & Hochreiter, S. (2017). Self-normalizing neural networks. In Advances in neural information processing systems (pp. 971-980).

45. Kingma, D. P., & Ba, J. (2014). Adam: A method for stochastic optimization. arXiv preprint arXiv:1412.6980.

46. D'Souza, M., Johnson, M., Dorn, J., Van Munster, C., Diederich, M., Kamm, C., ... & Walsh, L. (2018). Autoencoder-a new method for keeping data privacy when analyzing videos of patients with motor dysfunction (P4. 001).

47. Lintusaari, J., Vuollekoski, H., Kangasrääsiö, A., Skytén, K., Järvenpää, M., Marttinen, P., ... & Kaski, S. (2018). Elfi: Engine for likelihood-free inference. The Journal of Machine Learning Research, 19(1), 643-649.

48. Goldberger AL, Amaral LAN, Glass L, Hausdorff JM, Ivanov PCh, Mark RG, Mietus JE, Moody GB, Peng C-K, Stanley HE. PhysioBank, PhysioToolkit, and PhysioNet: Components of a New Research Resource for Complex Physiologic Signals (2003). Circulation. 101(23):e215-e220.

49. Ioffe, S., & Szegedy, C. (2015). Batch normalization: Accelerating deep network training by reducing internal covariate shift. arXiv preprint arXiv:1502.03167.

50. Sisson, S. A., Fan, Y., & Tanaka, M. M. (2007). Sequential monte carlo without likelihoods. Proceedings of the National Academy of Sciences, 104(6), 1760-1765.

51. Zhang, T., & Yang, X. (2018). G-SMOTE: A GMM-based synthetic minority oversampling technique for imbalanced learning. arXiv preprint arXiv:1810.10363.

52. Karabatsos, G., & Leisen, F. (2018). An approximate likelihood perspective on ABC methods. Statistics Surveys, 12, 66-104.

53. Meeds, E., Leenders, R., & Welling, M. (2015). Hamiltonian abc. arXiv preprint arXiv:1503.01916.